%% file: dt21.tex
\documentclass[journal]{IEEEtran}
\usepackage{cite}
\ifCLASSINFOpdf
  \usepackage[pdftex]{graphicx}
\else
\fi
\usepackage{amsmath}

\usepackage{multirow}
\usepackage{gensymb}
\usepackage{threeparttable}
\usepackage[caption=false]{subfig}
\usepackage{xcolor}
\newcommand{\ineq}[1]{\footnotesize$#1$\normalsize}{}
\newcommand{\mr}[1]{\textcolor{black}{#1}}
\newcommand{\minor}[1]{\textcolor{black}{#1}}
\definecolor{mygrey}{gray}{0.80}

\newcommand{\sm}{\text{{SpiNeMap}}}{}

\begin{document}
\bstctlcite{IEEEexample:BSTcontrol}
\title{On the Mitigation of Read Disturbances in Neuromorphic Inference Hardware}

\author{Ankita~Paul,
        Shihao~Song,
        Twisha~Titirsha~and
        Anup~Das% <-this % stops a space
\IEEEcompsocitemizethanks{\IEEEcompsocthanksitem A. Paul, S. Song, T. Titirsha, and A. Das, are with the Department
of Electrical and Computer Engineering, Drexel University,
PA, 19147.\protect\\

E-mail: \{ankita.paul,shihao.song,anup.das\}@drexel.edu}
\thanks{Manuscript received Month DD, Year; revised Month DD, Year.}}


\markboth{EEE Design \& Test,~Vol.~0, No.~0, Month~Year}%
{Titirsha \MakeLowercase{\textit{et al.}}: On the Mitigation of Read Disturbances in Neuromorphic Inference Hardware}



\maketitle

\begin{abstract}
\mr{
%Neuromorphic hardware are VLSI circuits that integrate synaptic memory closer to neural circuitry, mitigating the performance and energy bottleneck of conventional shared-memory systems in performing machine learning inference tasks. 
Non-Volatile Memory (NVM) cells are used in neuromorphic hardware to store model parameters, which are programmed as resistance states.
%(as resistance states) in such hardware.
NVMs suffer from the read disturb issue, where the programmed resistance state drifts upon repeated access of a cell during inference.
We show that resistance drifts 
%depends on 1) the criticality of a synaptic connection and 2) the number of synaptic activation 
can lower the inference accuracy. To address this, it is necessary to periodically reprogram model parameters to the hardware (a high overhead operation). % in order to maintain the desired accuracy.
We study read disturb failures of an NVM cell. Our analysis show both a strong dependency on model characteristics such as synaptic activation and criticality, and on the voltage used to read resistance states during inference.
We propose a system software framework to incorporate such dependencies in programming model parameters on NVM cells of a neuromorphic hardware.
At the core of our framework is a convex optimization formulation which aims to implement synaptic weights that have more activations and are critical, i.e., those that have high impact on accuracy on NVM cells that are exposed to lower voltages during inference. In this way, we increase the time interval between two consecutive reprogramming of model parameters.
We evaluate our system software with many emerging inference models on a neuromorphic hardware simulator and show a significant reduction in the system overhead.
}

\mr{
 }
% This article describes a software framework to evaluate read disturbance-related reliability issues in Non-Volatile Memory (NVM)-based neuromorphic hardware and quantify the system-level overhead in mitigating such issues when performing machine learning inference tasks.
\end{abstract}


% \fi
\IEEEpeerreviewmaketitle

\section{Introduction}
\input{sections/introduction}

\section{Resistance Drift Tolerance of Machine Learnig Workloads}\label{sec:accuracy}
\input{sections/accuracy}

\section{Resistance Drifts in Non-Volatile \\Memory Devices}\label{sec:device}
\input{sections/reliability}

\section{Workload Dependency of Inference Lifetime}\label{sec:workload}
\input{sections/workload}

\section{Proposed Design Methodology}\label{sec:methodology}
\input{sections/design_methodology}

% \section{Neuromorphic Computing}\label{sec:neuromorphic_computing}
% \input{sections/neuromorphic}

% \section{System-Level Approaches to Read Disturbance Management}\label{sec:system_level}
% \input{sections/system_level}

\section{Results and Discussion}\label{sec:evaluation}

\input{sections/evaluation}

\section{Conclusion}\label{sec:conclusion}
\input{sections/conclusions}

\section*{Acknowledgment}

\input{sections/ack}

\bibliographystyle{IEEEtran}
\bibliography{commands,disco,external}

\input{sections/bio}

\end{document}

%% file: sections/introduction.tex
\IEEEPARstart{N}{euromorphic} systems are integrated circuits designed to mimic the neural architecture in primates. 
\mr{Here, neural circuity is tightly coupled with synaptic storage, which eliminates the performance and energy bottlenecks of shared-memory systems for machine learning inference~\cite{sentryos}.
Non-Volatile Memory (NVM) cells such as oxide-based resistive switching random access memory (OxRRAM) can implement multilevel analog operations, which make them ideal candidates for storing model parameters, i.e., the synaptic weights of a machine learning model~\cite{mallik2017design}.
}

% are used to implement the synaptic weights in a neuromorphic hardware due to their low power consumption, CMOS-compatible scaling, and multilevel analog operations~\cite{mallik2017design}.
% NVM-based neuromorphic systems facilitate energy-efficient implementation of machine learning inference tasks on energy-constrained environments such as Embedded Systems and IoT Edge devices.

\mr{
For use as an inference hardware, trained model parameters are programmed as resistance states on OxRRAM cells of the hardware. Once programmed, the hardware is expected to perform inference continuously, without having to reprogram the model parameters.
}
Unfortunately, OxRRAM cells suffer from the read disturb issue, where a cell's \mr{resistance state may drift from its programmed value upon repeated access during inference~\cite{shim2020impact}. We show that resistance drifts can lead to a lower inference accuracy (see Section~\ref{sec:evaluation}).
%(Fig.~\ref{fig:drop}), such resistance drifts can lead to a significant reduction in the inference accuracy.
}

\mr{
One system-level technique to mitigate read disturbances in a neuromorphic hardware is to periodically reprogram the trained parameters to the OxRRAM cells of the hardware.
Reprogramming of model parameters involves transferring the synaptic weights from the main memory (primary storage) to the neuromorphic hardware via bandwidth-limited memory channels (see Fig.~\ref{fig:overview}). Additionally, NVM cells require the long-latency program-and-verify (P\&V) scheme to configure their resistance states~\cite{milo2019multilevel}. These factors increase the time it takes to reprogram model parameters on OxRRAM cells.
When a model is being reprogrammed, the hardware is unavailable to perform inference operations. Therefore, the performance overhead associated with periodic reprogramming is
\begin{equation}
    \label{eq:reprogramming_overhead}
    \footnotesize \text{reprogram overhead} = \frac{tRPT}{tRPI},
\end{equation}
where \ineq{tRPT} defines the reprogramming time of the model and \ineq{tRPI} defines the interval at which the model is being reprogrammed to the hardware.
}

\mr{
We show that periodic reprogramming leads to a high system overhead even for smaller models like LeNet and AlexNet, and is expected to become a critical performance bottleneck for emerging large models such as VGGNet, ResNet, and DenseNet.
Our \textbf{objective} is to minimize this overhead by increasing the reprogram interval \ineq{tRPI}. To this end, we make the following three key observations.
}

\mr{
\textbf{\textit{Observation 1:}} \textit{Different synaptic connections of a machine learning model have different tolerance to resistance drift and they impact model accuracy differently. See Sec.~\ref{sec:accuracy}.}
}

\mr{
\textbf{\textit{Observation 2:}} \textit{OxRRAM cells in a neuromorphic hardware exhibit variation in read disturbance due to a difference in the exposed read voltage used during inference. See Sec.~\ref{sec:device}.}
}

\mr{
\textbf{\textit{Observation 3:}} \textit{Activation of a synaptic connection in a model is workload-dependent and it leads to a difference in the amount of resistance drift within the model. See Sec.~\ref{sec:workload}.}
}

\mr{
Based on these three observations, we propose a system software framework that incorporates this application and voltage-dependent characteristics of read disturbance of OxRRAM cells in implementing a machine learning model on the hardware. The \textbf{key idea} is to implement the synaptic weight of connections that have higher activations and lead to higher accuracy drop on NVM cells that are exposed to lower voltages during inference. In this way, we are able to sustain larger resistance drifts of synaptic weights before reprogramming of model parameters on OxRRAM cells becomes necessary.
}
% \mr{
% \textbf{\textit{Observation 2:}} \textit{Read disturbance of an OxRRAM cell is a function of the voltage used to read its resistance state.}
% }

% \mr{
% \textbf{\textit{Observation 3:}} \textit{Read voltage of OxRRAM cells varies widely within a neuromorphic hardware.}
% }

%To estimate this performance overhead, let \ineq{tRPT} defines the reprogramming time of a model and \ineq{tRPI} defines the reprogramming interval. 

\mr{
A preliminary version of this system software framework is proposed in our prior work~\cite{song2021improving}. Here we extend this framework in four key directions -- 1) introducing overhead due to reprogramming of model parameters as a key performance metric, 2) extending the system software framework to periodically reprogram model parameters to a neuromorphic hardware in order to maintain integrity of machine learning tasks,  3) a convex optimization formulation of cluster mapping to crossbar in order to reduce the system overhead, and 4) exploiting machine learning model characteristics to identify non-critical model parameters and eliminating them from the critical path of deciding the reprogramming interval. In this way, our convex optimizer is able to increase the reprogramming interval compared to~\cite{song2021improving}, thereby significantly reducing the system overhead (see Section~\ref{sec:evaluation}).
}

\mr{
We integrate the proposed system software framework inside NeuroXplorer~\cite{neuroxplorer}, a cycle-accurate simulator of neuromorphic hardware and evaluate it using five commonly-used machine learning inference applications. Results demonstrate an average 35\% reduction of system overhead.
}

%% file: sections/accuracy.tex
\begin{figure}[h!]
	\begin{center}
		\vspace{-10pt}
		\includegraphics[width=0.99\columnwidth]{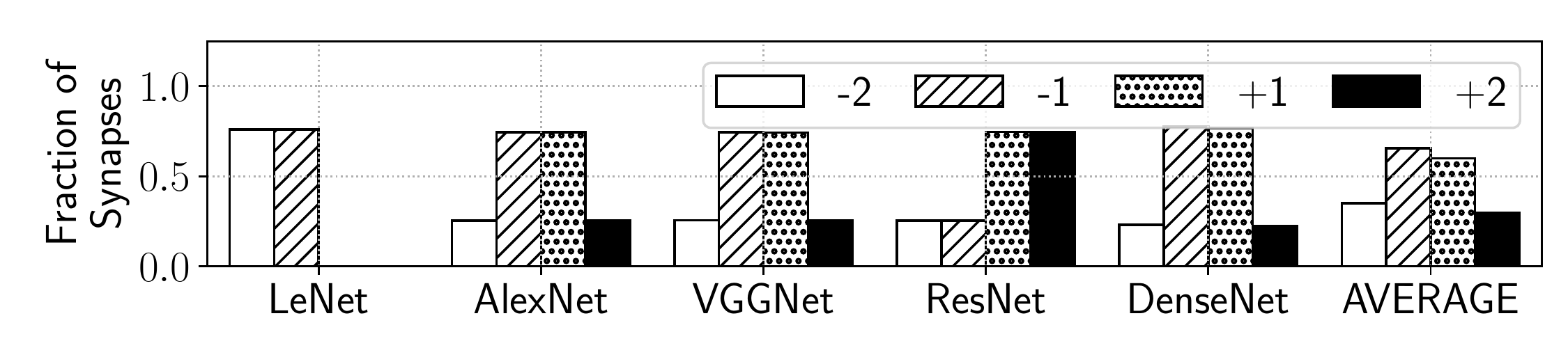}
		\vspace{-10pt}
		\caption{\mr{Fraction of total synapses in the fully-connected layer that leads to accuracy drop due to resistance drift.}}
		%\vspace{-10pt}
		\label{fig:accuracy}
		\vspace{-10pt}
	\end{center}
\end{figure}

\mr{
Synaptic connections of a machine learning workload have varying tolerances to resistance drift. This impacts accuracy differently. To illustrate this, we consider 2-bit quantized versions of five commonly-used convolutional neural networks (CNNs) -- LeNet (1989), AlexNet (2012), VGGNet (2015), ResNet (2015), and DenseNet (2017). There are \minor{three weight levels used in these models, corresponding to ternary values of -1, 0, and +1~\cite{han2015deep}.} Figure~\ref{fig:accuracy} illustrates the fraction of total synapses in the fully-connected layer that leads to 1\% or higher accuracy drop. We report results for the following four configurations -- 1) resistance reduction by two levels (``-2"), 2) resistance reduction by one level (``-1"), 3) resistance increase by 1 level (``+1"), and 4) resistance increase by two levels (``+2"). \footnote{\minor{We note that if a synaptic weight is +1, then the synapse is tolerant to resistance drifts in the positive direction. Similarly, if a synaptic weight is -1, then the synapse is tolerant to drifts in the negative direction. Such cases are included in the results of Figure~\ref{fig:accuracy}.}}
We make the following \minor{three} key observations.
}

\mr{
First, synapses in a machine learning model have varying tolerance to resistance drift. On average, only 35\% of synapses show accuracy drop when their resistance drifts by two levels in the negative direction, only 66\% when resistance drifts by one level in the negative direction, only 60\% when resistance drifts by one level in the positive direction, and only 30\% when resistance drifts by two levels in the positive direction. The reason for such variations is two-fold. First, most machine learning models are over-parameterized. Therefore, resistance levels of non-critical synapses do not impact the accuracy. Second, due to the approximate training using the backpropagation algorithm, a drift in the resistance level of some synapses may not impact accuracy significantly. {To this end, we note that the synaptic weight value of a non-critical synapse may not necessarily be close to zero. It simply means that any change of its weight value may not impact accuracy. For this reason, any neuron and synapse pruning strategy such as~\cite{han2015deep} will not eliminate non-critical synapses that are non-zero.}
}

\minor{
Second, for LeNet, only a small fraction (less than 1\%) of synapses lead to accuracy drop when resistance drifts by +1 and +2. This is because most of synaptic weights of LeNet are positive. So any transition in the positive direction results in no significant accuracy impact. 
}

\mr{
Third, tolerance to resistance drift depends on the specific CNN model and therefore, model-specific solutions are needed. Our proposed approach is the following. First, we identify the critical synapses, i.e., those that have high impact on accuracy by analyzing a CNN model. Next, we exploit device characteristics and mapping alternatives to minimize the negative impact of resistance drift.
}

\mr{
To motivate our solution, we discuss resistance drift in NVMs, focusing on OxRRAM devices.
}

%% file: sections/reliability.tex
%There are many sources of reliability issues in OxRRAM-based neuromorphcic hardware. 
% Examples include aging of CMOS devices in neuron circuits due to high voltage operations of OxRRAM cells~\cite{reneu}, %crosstalk between neighboring OxRRAM cells due to high integration density, 
% limited programmability of OxRRAM cells due to irreversible wear-out~\cite{espine}, and read disturbances caused by temporary structural alteration in OxRRAM cells due to passage of current~\cite{song2021improving}. These reliability issues can be mitigated using optimization techniques at the device-, circuit-, and system-level.

% %We focus on system-level reliability improvement techniques. To this end, we have previously proposed RENEU~\cite{reneu} and NCRTM~\cite{ncrtm} for aging improvement, and eSpine~\cite{espine} for write endurance improvement. In these approaches, the system software explores different implementation alternatives and selects the one that satisfies a given reliability objective.
%This work addresses the read disturbance-related reliability issues in OxRRAM-based neuromorphic hardware. 
%To study resistance drift related reliability issues, we briefly review the OxRRAM technology.

\subsection{Oxide-based Resistive RAM (OxRRAM) Technology}
\begin{figure}[h!]
	\begin{center}
		\vspace{-10pt}
		\includegraphics[width=0.69\columnwidth]{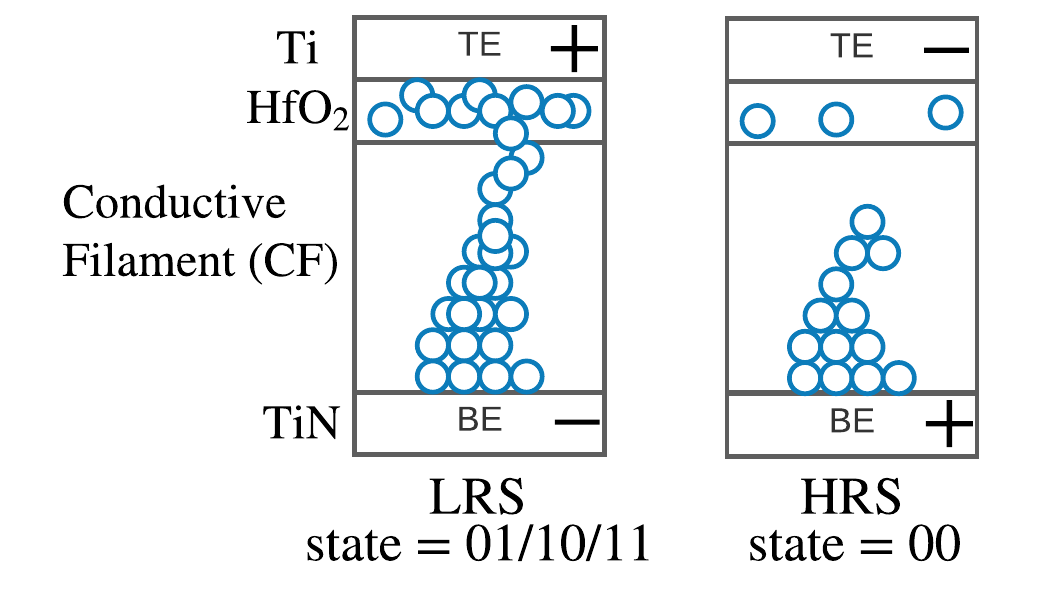}
		\vspace{-10pt}
		\caption{Operation of an OxRRAM cell with the $\text{HfO}_2$ layer sandwiched between the metals Ti (top electrode) and TiN (bottom electrode). The left subfigure shows the formation of LRS states with the formation of conducting filament (CF). %This represents logic states 01, 10, and 11. 
		The right subfigure shows the depletion of CF on application of a negative voltage on the TE. 
		%This represents the HRS state or logic 00.
		}
		%\vspace{-10pt}
		\label{fig:RRAM}
		\vspace{-10pt}
	\end{center}
\end{figure}

The resistance switching random access memory (OxRRAM) technology presents an attractive option for implementing the synaptic cells of a crossbar due to its demonstrated potential for low-power multilevel operation and high integration density~\cite{mallik2017design}. An OxRRAM cell is composed of an insulating film sandwiched between conducting electrodes forming a metal-insulator-metal (MIM) structure (see Figure~\ref{fig:RRAM}). Recently, filament-based metal-oxide OxRRAM implemented with transition-metal-oxides such as HfO${}_2$, ZrO${}_2$, and TiO${}_2$ has received considerable attention due to their low-power and CMOS-compatible scaling.

Synaptic weights are represented as conductance of the insulating layer within each OxRRAM cell. To program an OxRRAM cell, elevated voltages are applied at the top and bottom electrodes, which re-arranges the atomic structure of the insulating layer. Figure~\ref{fig:RRAM} shows the High-Resistance State (HRS) and the Low-Resistance State (LRS) of an OxRRAM cell. An OxRRAM cell can also be programmed into intermediate low-resistance states, allowing its multilevel operations. 
%We consider each OxRRAM cell to be programmed to one HRS and three LRS states, implementing two bits per synapse.

\subsection{Read Disturbance Issues of OxRRAM Cells}
\begin{figure}[h!]
	\centering
	\vspace{-5pt}
	\centerline{\includegraphics[width=0.89\columnwidth]{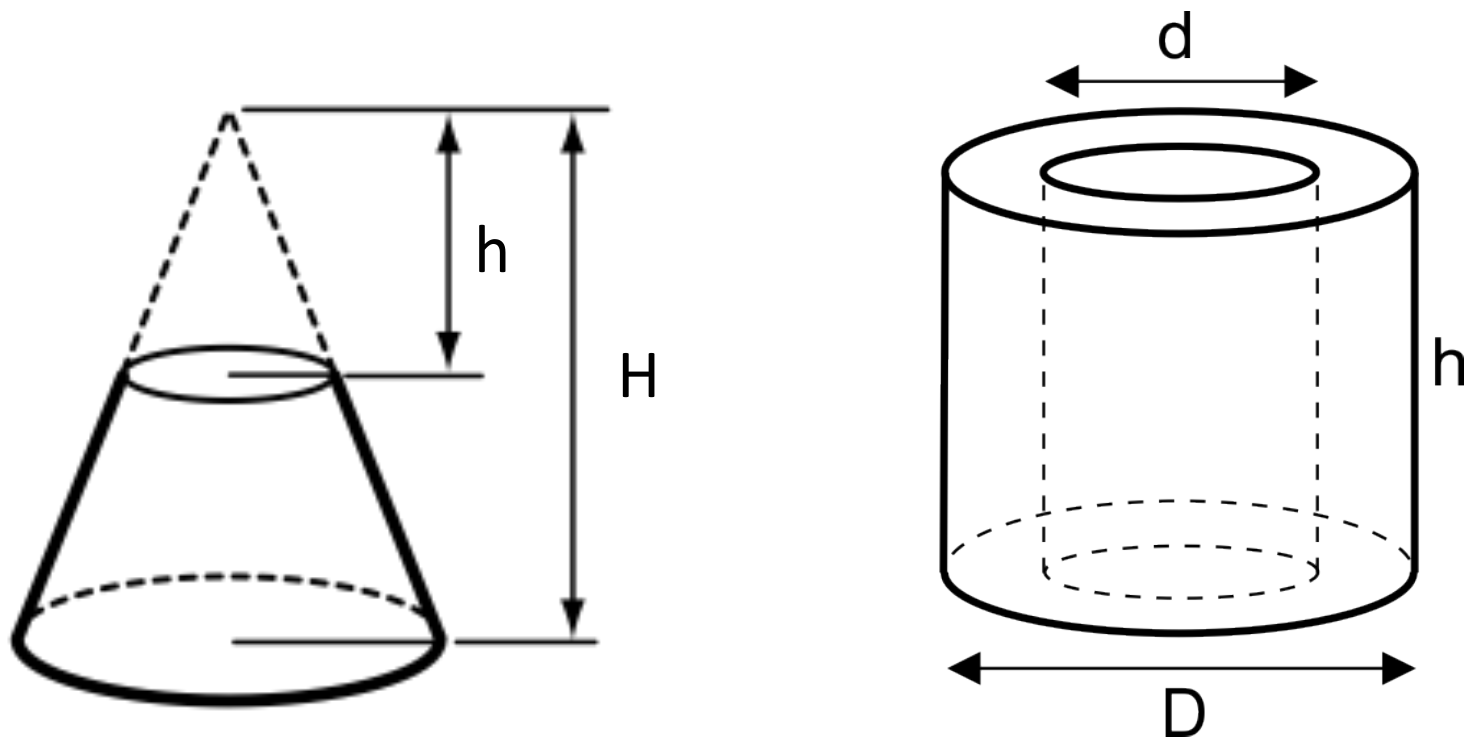}}
	\vspace{-10pt}
	%\caption{An example of spiking neural network.}
	\caption{\minor{Read disturbances due to structural alteration in an OxRRAM cell. The left subfigure shows a reduction of the conductive filament gap (i.e., read disturbance of HRS state) on the application of a stress voltage. The right subfigure shows the lateral growth of the conductive filament (i.e., read disturbance of LRS state) due to application of a stress voltage.}}
	\vspace{-5pt}
	\label{fig:read_disturbances}
\end{figure}

In OxRRAM technology, \mr{the transition from HRS to one of the LRS states} is governed by a sudden decrease of the vertical filament gap on application of a stress voltage during spike propagation~\cite{shim2020impact}.
This is illustrated in the left subfigure of Figure~\ref{fig:read_disturbances} where the vertical filament gap is shown to reduce by an amount $h$. This may result in a conducting filament between the two metal layers causing the resistive state to change from HRS to LRS.
\mr{
The rate of change of the filament gap of the OxRRAM cell 
%at the \ineq{(i,j)^\text{th}} location in the crossbar 
is
\begin{equation}
    \label{eq:hrs}
    %\vspace{-10pt}
    \footnotesize \frac{dg}{dt} = -\vartheta_0\cdot e^{-\frac{E_a}{kT}}sinh\left(\frac{\gamma\cdot a_0}{L}\cdot\frac{qV}{kT}\right) \text{, where } \gamma = \gamma_0 - \beta\cdot\frac{g}{g_0}^3
\end{equation}
In the above equation, \ineq{t} defines the state transition time, \ineq{g_0} is the initial filament gap of the OxRRAM cell, \ineq{V} is the voltage applied to the cell, \ineq{\gamma} is the local field enhancement factor, which is related to the gap \ineq{g}, \ineq{a_0} is the atomic hoping distance, \minor{\ineq{E_a} is the activation energy, \ineq{k} is the Boltzmann constant, \ineq{T} is the temperature (in Kelvin), \ineq{L} is the length of the vertical filament, \ineq{q} is the filament charge, \ineq{\vartheta_0} is a constant related to vertical filament growth, and \ineq{\gamma_0 \text{ and } \beta} are fitting constants.
}}

The transition from one of the LRS states is governed by the lateral filament growth~\cite{shim2020impact}. 
This is illustrated in the right subfigure of Figure~\ref{fig:read_disturbances}.
The time for state transition in the OxRRAM cell is given by 
\begin{equation}
    \label{eq:lrs}
    \footnotesize t (LRS) = 10^{-14.7\cdot V + 6.7} \text{sec}
\end{equation}

% Using Equations~\ref{eq:hrs} and \ref{eq:lrs}, the read endurance of the \ineq{(i,j)^\text{th}} RRAM cell can be derived as 
% \begin{equation}
%     \label{eq:endurance}
%     \footnotesize E_{i,j} (HRS/LRS) = \frac{t_{i,j} (HRS/LRS)}{1~ms \text{ (spike duration)}}
% \end{equation}

If the state transition time of an OxRRAM cell is 1000~ms, then a single quasi-static read operating using one 1000 ms read pulse or equivalently, 1000 read accesses using 1-ms spike pulses can lead to an abrupt change in the cell's state.
\footnote{Apart from resistance drift, there are also other forms of reliability issues reported for OxRRAM in the context of neuromorphic hardware~\cite{chaudhuri2019hardware,spyrou2021neuron,espine,ankita_igsc}.}

\begin{figure}[h!]
	\centering
	\vspace{-5pt}
	\centerline{\includegraphics[width=1.09\columnwidth]{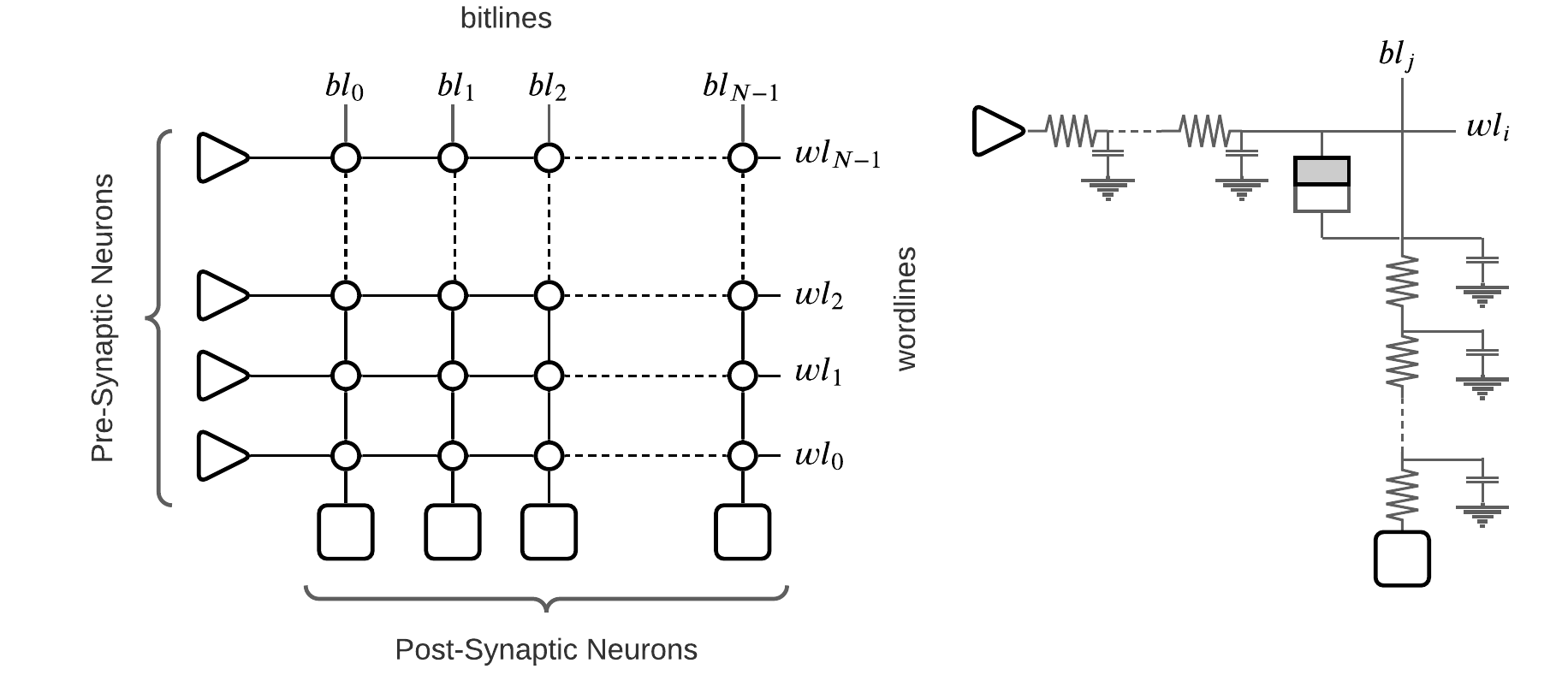}}
	\vspace{-10pt}
	%\caption{An example of spiking neural network.}
	\caption{An $N \times N$ crossbar showing the parasitic components within.}
	\vspace{-5pt}
	\label{fig:crossbar}
\end{figure}

\mr{
From Equations~\ref{eq:lrs} \& \ref{eq:hrs}, we see that the state transition time of an OxRRAM cell depends heavily on the voltage of operation of the cell. To this end, we investigate the internal architecture of a processing core in a neuromorphic hardware. In many recent designs, analog crossbars are used as cores. Figure~\ref{fig:crossbar} (left) shows an \ineq{N\times N} crossbar where the OxRRAM cells are organized in a two-dimensional grid with horizontal wordlines and vertical bitlines. Pre-synaptic neurons are mapped along wordlines and post-synaptic neurons along bitlines as shown in the figure. The synaptic weight between a pre-synaptic neuron (\ineq{n_i}, placed on the \ineq{i^\text{th}} wordline) and a post-synaptic neuron (\ineq{n_j}, placed on the \ineq{j^\text{th}} bitline) is programmed as conductance of the OxRRAM cell (\ineq{i,j}) located at the intersection of \ineq{i^\text{th}} wordline and \ineq{j^\text{th}} bitline.
}

\mr{
For forward propagation of neuron excitation, a spike voltage is created by \ineq{n_i}, which generates a current that propagates to the neuron \ineq{n_j} via the conductance of the \ineq{(i,j)^\text{th}} OxRRAM cell. Figure~\ref{fig:crossbar} (right) shows the parasitic components on such current paths.
}
Formally, the number of parasitic components on the current path via the $(i,j)^\text{th}$ OxRRAM cell is $(i+j+1)$.

Parasitic components on bitlines and wordlines of a crossbar create variation in currents propagating via different OxRRAM cells of the crossbar; higher the number of parasitic components, smaller is the current, and vice versa. Therefore, the current through \ineq{(0,0)^\text{th}} OxRRAM cell is higher than \ineq{(N-1,N-1)^\text{th}} OxRRAM cell in an \ineq{N\times N} crossbar.

\begin{figure}[h!]
	\centering
	%\vspace{-10pt}
	\centerline{\includegraphics[width=0.99\columnwidth]{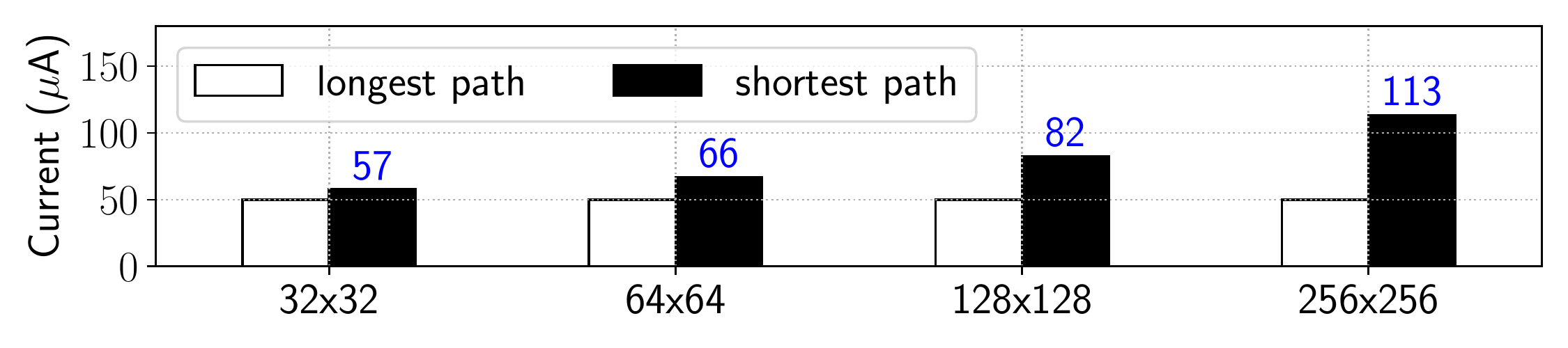}}
	%\vspace{-10pt}
	\caption{\mr{Difference between current on the shortest and longest paths in a crossbar for different crossbar sizes.}}
	%\vspace{-10pt}
	\label{fig:current_crossbar_size}
\end{figure}

\mr{
Figure~\ref{fig:current_crossbar_size} shows the difference between currents on the shortest and longest path for 32x32, 64x64, 128x128, and 256x256 crossbars at {65nm} process node. The input spike voltage of the pre-synaptic neurons is set to generate \ineq{50\mu A} on the longest path. This current value corresponds to the current needed to read the resistance state of the OxRRAM cell on this path.
We observe that the current on the longest path is lower than the shortest path by 13.3\% for 32x32, 25.1\% for 64x64, 39.2\% for 128x128, and 55.8\% for 256x256 crossbar.
}

\begin{figure}[h!]%
    \centering
    \subfloat[Voltage variation in a 128x128 crossbar at 65nm node.\label{fig:voltage_variation}]{{\includegraphics[width=4.0cm]{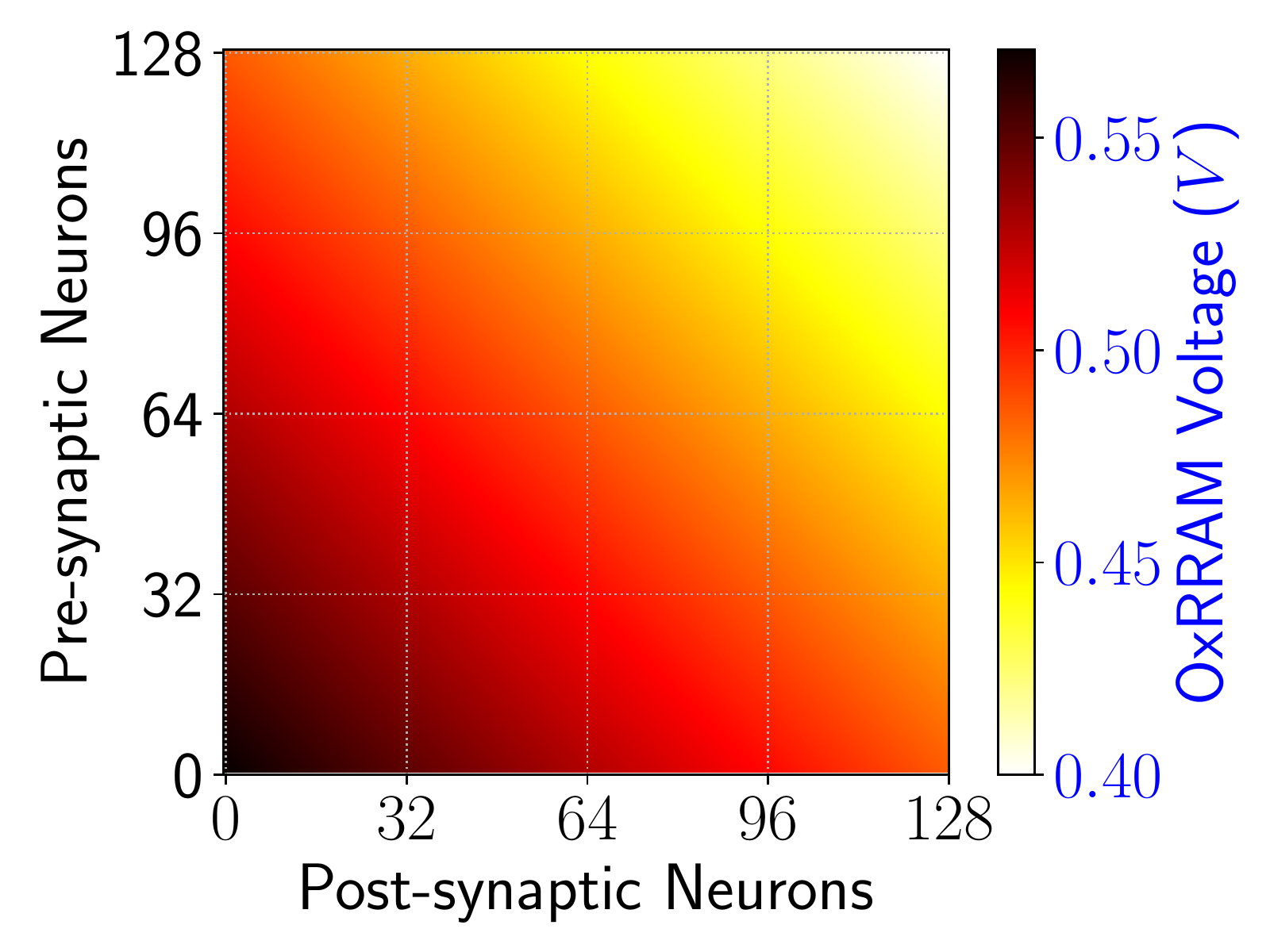} }}%
    \quad
    \subfloat[State transition times in a  128x128 crossbar with all cells programmed to the HRS state.\label{fig:hrs_var}]{{\includegraphics[width=4.0cm]{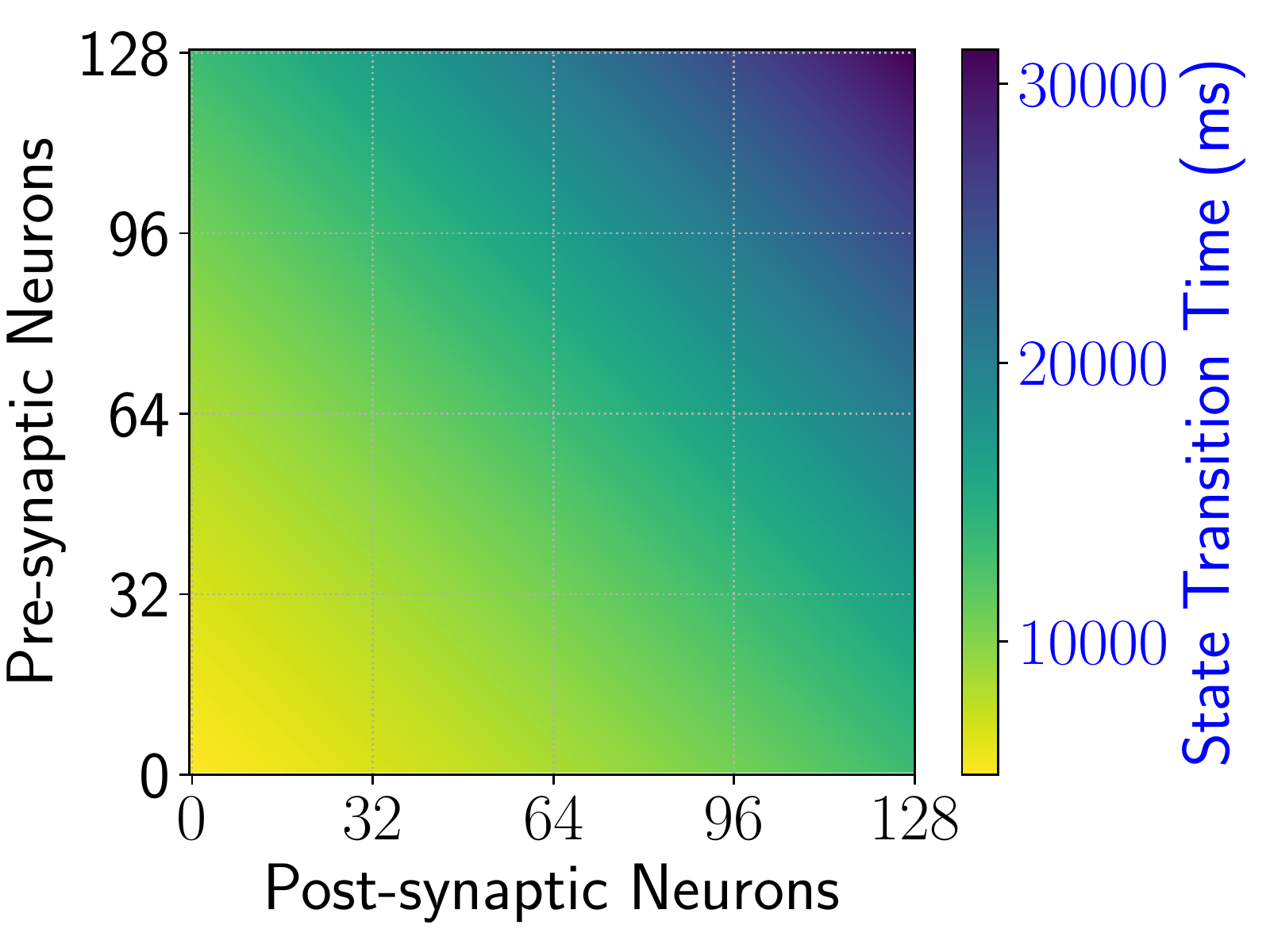} }}%
    %\vspace{-5pt}
    \caption{Variation of voltage and state transition time in a 128x128 crossbar.}%
    \label{fig:current_endurance_map}%
\end{figure}

\mr{
Current variation in a crossbar leads to a difference in the voltage applied across different OxRRAM cells in a crossbar. This is illustrated in Figure~\ref{fig:voltage_variation}, where the minimum and maximum voltages are 0.4 V and 0.57 V, respectively. Such voltage differences cause variation of the state transition time (see Equations~\ref{eq:hrs} \& \ref{eq:lrs}). Figure~\ref{fig:hrs_var} shows such variation for OxRRAM cells in the crossbar, with each cell programmed to the HRS state. The minimum and maximum state transition times are 5,227 ms and 31,214 ms, respectively.
%Endurance variation for LRS states can also be obtained using our circuit simulation framework (see Section~\ref{sec:evaluation}).
}

\begin{figure}[h!]
	\centering
	\vspace{-5pt}
	\centerline{\includegraphics[width=0.99\columnwidth]{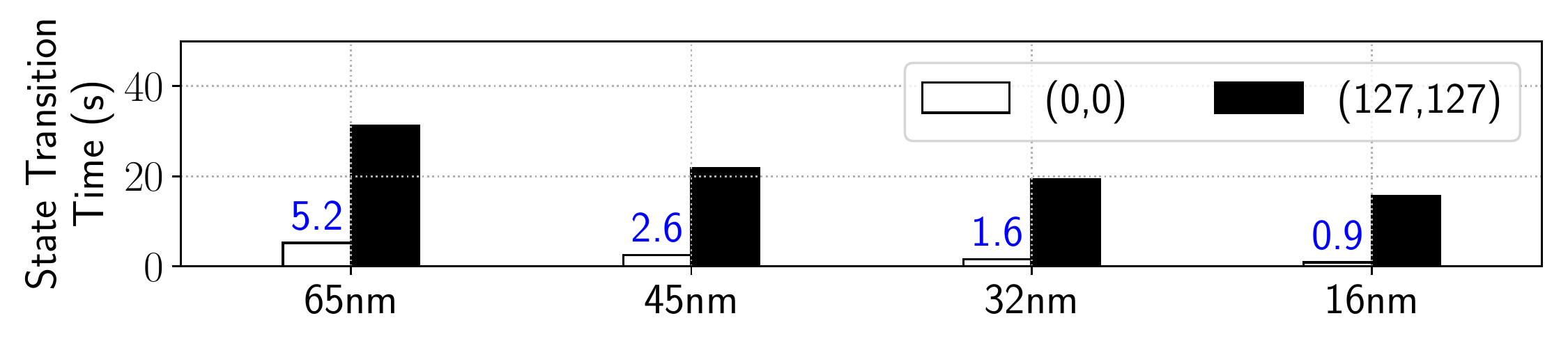}}
	\vspace{-10pt}
	%\caption{An example of spiking neural network.}
	\caption{Variation in state transition time of OxRRAM cells in a crossbar as a function of current.}
	\vspace{-5pt}
	\label{fig:read_endurance_variations}
\end{figure}

\mr{
In a recent work, we have shown that the unit parasitic resistance of bitlines/wordlines increases from \ineq{1\Omega} at 65~nm to \ineq{3.8\Omega} at 16~nm~\cite{espine}. Such increase in the value of parasitic resistance leads to a higher voltage applied across each OxRRAM cell in the crossbar, which further reduces its state transition time. To illustrate this,
}
Figure~\ref{fig:read_endurance_variations} shows the variation in state transition time of OxRRAM cells in a crossbar at different process technology nodes.
% \begin{figure}[h!]
% 	\centering
% 	%\vspace{-10pt}
% 	\centerline{\includegraphics[width=0.49\columnwidth]{images/current_map_50.65.pdf}}
% 	%\vspace{-10pt}
% 	\caption{\mr{Voltage variation in a 128x128 crossbar at 65nm node.}}
% 	%\vspace{-10pt}
% 	\label{fig:voltage_variation}
% \end{figure}
% Figure~\ref{fig:read_endurance_variations} shows the variation in state transition time of OxRRAM cells in a crossbar, where each cell is programmed to the HRS state. The fitting parameters are adjusted to achieve a 10s transition time of the $(0,0)^\text{th}$ cell in the crossbar.
We make the following two key observations. First, the state transition time decreases with technology scaling. \mr{This is due to an increase in the voltage within the crossbar at scaled nodes.} Second, the variation of state transition time increases at smaller nodes due to higher voltage and current variations~\cite{espine}.

\begin{figure}[h!]
	\centering
	\vspace{-5pt}
	\centerline{\includegraphics[width=0.99\columnwidth]{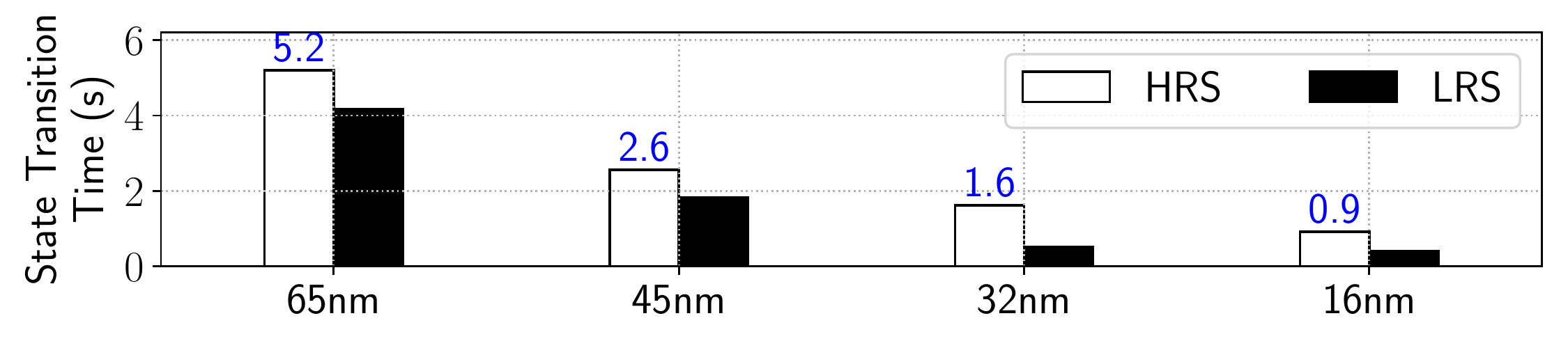}}
	\vspace{-10pt}
	%\caption{An example of spiking neural network.}
	\caption{Variation in state transition time of OxRRAM cells in a crossbar as a function of the resistance state.}
	\vspace{-5pt}
	\label{fig:state_dependence}
\end{figure}

Finally, the state transition time of OxRRAM cells also depends on the resistance state. Figure~\ref{fig:state_dependence} shows the dependence of the minimum state transition time of  OxRRAM cells in a crossbar for the four process technology nodes. We make two key observations. First, the state transition time reduces with technology scaling, which we have analyzed before. Second, the state transition time of an OxRRAM cell is higher when the cell is programmed in the HRS state for all process technology nodes. This is because the vertical filament growth phenomena (in HRS state) in OxRRAM technology is slower than the lateral filament growth (in LRS state).

\mr{
During each inference operation, OxRRAM cells of a crossbar propagate spikes from a machine learning workload. To compute the \textbf{inference lifetime} of an OxRRAM cell, which is defined as the number of inference operations it takes for the resistance state of the cell to drift from its programmed value, we let \ineq{\eta} be the average number of spikes through the cell per inference operation. Formally, inference lifetime \ineq{\mathcal{L}} is
\begin{equation}
    \label{eq:inference_lifetime}
    \footnotesize \mathcal{L} = \frac{t(LRS/HRS)}{\eta}
\end{equation}
}

\mr{
To ensure integrity of machine learning, i.e., to prevent accuracy drop, the OxRRAM cell must be reprogrammed to the original resistance state once every inference lifetime. Since different OxRRAM cells in a neuromorphic crossbar have different inference lifetime, the reprogramming interval \ineq{tRPI} of model parameters to the hardware (see Eq.~\ref{eq:reprogramming_overhead}) is defined as the minimum inference lifetime of all OxRRAM cells in the crossbar, i.e.,
\begin{equation}
    \label{eq:reproggraing_interval}
    \footnotesize tRPI = \underset{\forall i,j}{\texttt{min}} ~\mathcal{L}_{i,j}
\end{equation}
}

\mr{The number of spikes propagating through an OxRRAM cell depends on the machine learning workload and how the workload is mapped to the crossbar. This is described next.}

% During inference, OxRRAM cells in a crossbar propagates different number of spikes. We define \textbf{transition lifetime} of an OxRRAM cell as the minimum number of spikes that can trigger a state transition in the cell, Formally,
% \begin{equation}
%     \label{eq:transition_capacity}
%     \footnotesize L_{i,j} = \frac{t_{i,j} (LRS/HRS)}{\tau_s},
% \end{equation}
% where \ineq{t_{i,j}(LRS/HRS)} is the state transition time of the cell and \ineq{\tau_s} is the spike duration. The \textbf{transition capacity} defines the minimum number of times an OxRRAM cell needs to be reprogrammed during an inference operation to prevent loss of its resistance state, i.e.,
% \begin{equation}
%     \label{eq:reprogramming_count}
%     \footnotesize C_{i,j} = \frac{S_{i,j}}{L_{i,j}},
% \end{equation}
% where \ineq{S_{i,j}} is the number of spikes propagating through the cell during an inference operation.

% Using the observations of Figures~\ref{fig:read_endurance_variations} and \ref{fig:state_dependence}, we \textbf{conclude} that the transition capacity of an OxRRAM cell in a crossbar depends on 1) the resistance state programmed on the cell (i.e., state dependency), 2) the placement of the cell in the crossbar (i.e., current dependency), and 3) the number of spikes propagating through the cell for a given inference operation (i.e., input dependency).

%% file: sections/workload.tex
\mr{
To understand the workload dependency of inference lifetime, we focus on Equation~\ref{eq:inference_lifetime}. \minor{Here, \ineq{\eta} is the average spikes per image through an OxRRAM cell implementing the machine learning workload. This is computed as follows. Consider our machine learning model is represented as \ineq{\mathcal{M}(N,S)} with the set \ineq{N} of neurons and the set \ineq{S} of synapses. 
%Since information in SNNs are encoded in terms of spikes, when an image is presented to the model for inference each synapse \ineq{s_i\in S} will communicate a certain number of spikes from its pre-synaptic neuron to its post-synaptic neuron. This number depends on the input image presented to the model and the information coding used (e.g., rate coding vs inter-spike interval coding). 
If \ineq{x_i} is the number of spikes communicated via \ineq{s_i} from its pre-synaptic to its post-synaptic neuron, then the total number of spikes for the image is \ineq{\sum_i x_i}. In our implementation, each synapse (i.e., its weight) is programmed on an individual OxRRAM cell. Therefore, the number of spikes through all OxRRAM cells of the hardware is \ineq{\sum_i x_i}. We compute the average number of spikes per image through an individual OxRRAM cell as the sum of spikes for all images inferred by the model averaged over the number of images and synapses, i.e.,
\begin{equation}
    \label{eq:average_spike_per_image}
    \footnotesize \text{Avg. Spike Per Image} = \frac{\sum_{j=1}^I \sum_i x_i}{I\times |S|},
\end{equation}
where \ineq{I} is the number of images inferred by \ineq{\mathcal{M}}.
}
}

\begin{figure}[h!]
	\begin{center}
		\vspace{-10pt}
		\includegraphics[width=0.99\columnwidth]{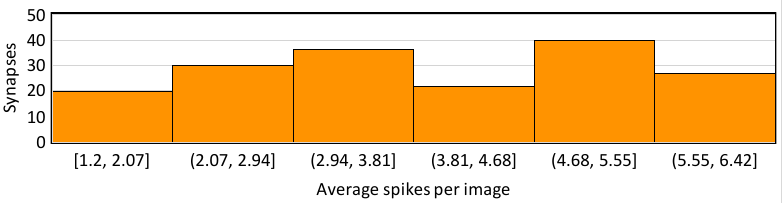}
		\vspace{-20pt}
		\caption{Spike distribution across the synapses of VGGNet.}
		%\vspace{-10pt}
		\label{fig:vgg_distribution}
		\vspace{-15pt}
	\end{center}
\end{figure}

\mr{Figure~\ref{fig:vgg_distribution} shows the histogram of average spikes per image propagating through the synapses of VGGNet. We collected these statistics by analyzing CIFAR-10 training and test datasets. \minor{We see that there are 20 synapses in the model that communicate between 1-2 spikes per image, 30 synapses that communicate between 2-3 spikes per image, and so on. Therefore, some synapses propagate more spikes than others.% when inferring an image.}
}
}

\mr{
If we consider two different synapses of a model with different spike count, then the one with a higher number of spikes will result in a lower inference lifetime when mapped to the OxRRAM cell at a specific position in the crossbar. 
}

\begin{figure}[h!]
 	\centering
    \vspace{-6pt}
 	\centerline{\includegraphics[width=0.99\columnwidth]{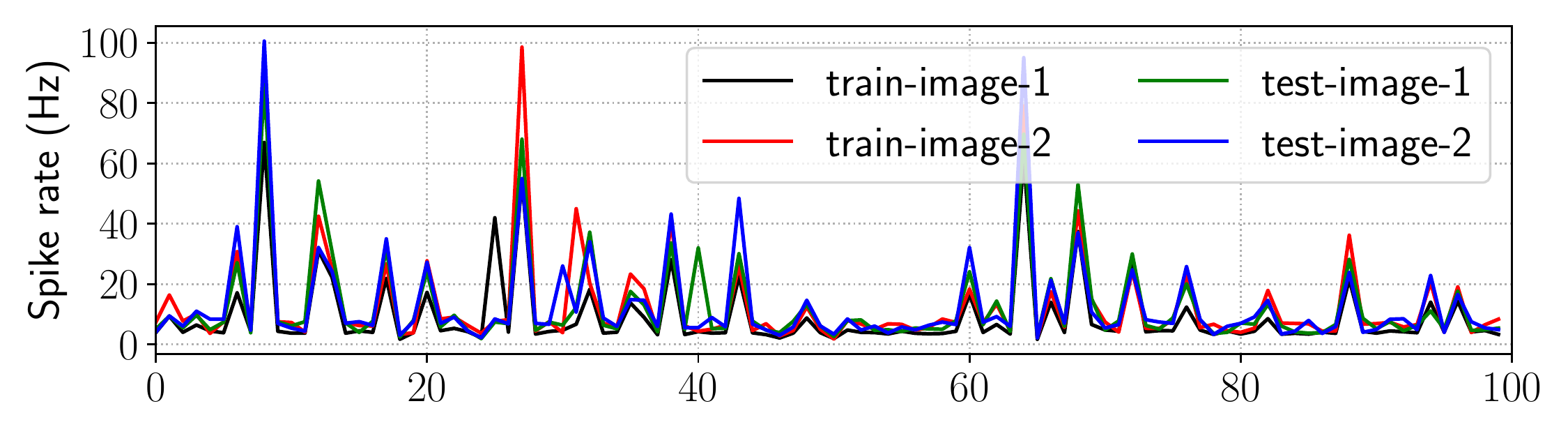}}
 	\vspace{-10pt}
 	\caption{Spike rate of 100 randomly-selected neurons in VGGNet for 2 training images and 2 test images.}
    %\vspace{-10pt}
 	\label{fig:vggnet}
\end{figure}

\mr{
Additionally, the spike count on a synapse also depends on the input presented to a model. To illustrate this, Figure~\ref{fig:vggnet} plots the spike firing rate of 100 randomly-selected neurons in VGGNet. We observe that the spike firing rate of a neuron in VGGNet depends on the image presented to the model.
}

\mr{
The proposed design methodology incorporates such application and model-dependent behavior to better optimize the synapse mapping to OxRRAM cells. This is described next.\footnote{The scope of the current work is on design-time approaches in mitigating resistance drift. Our future work will involve designing a run-time framework to evaluate spike count of synapses based on the model input and enable remapping of the synaptic connections to further reduce
the system overhead.}
}

%% file: sections/design_methodology.tex
\begin{figure}[h!]
	\centering
	%\vspace{-10pt}
	\centerline{\includegraphics[width=1.02\columnwidth]{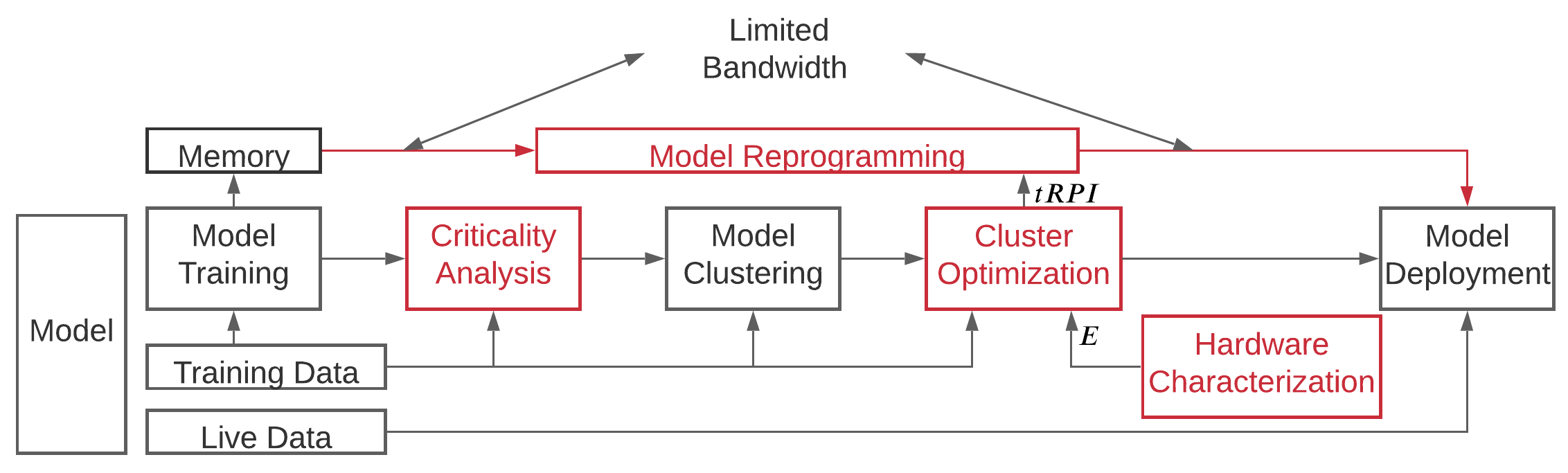}}
	%\vspace{-10pt}
	\caption{\mr{System architecture.}}
	%\vspace{-10pt}
	\label{fig:overview}
\end{figure}

\mr{
Figure~\ref{fig:overview} shows the proposed system architecture designed in NeuroXplorer~\cite{neuroxplorer}. A machine learning model is first trained using training data. The model parameters are stored in memory. The trained model is clustered using the graph partitioning algorithm of NeuroXplorer. For each cluster, an optimization is performed to map the neurons and synapses of the cluster to the OxRRAM cells of a crossbar, by exploiting 1) spike data collected from the training set and 2) technology-specific state-transition time data obtained from characterizing the hardware. The cluster optimization step generates the parameter reprogramming interval \ineq{tRPI}, which is then used to periodically reprogram the model parameters to the hardware via bandwidth-limited memory channels. The new blocks that we introduce in NeuroXplorer are shown in red in Fig.~\ref{fig:overview}. 
}

\mr{
We introduce the following notations to formulate the cluster optimization problem.
}

\mr{
% \begin{table}[t]
% %\vspace{-10pt}
% %\begin{small}
% \colorbox{white}{
% %\caption{Legends}
% \begin{minipage}[b]{0.99\linewidth}
% %\begin{small}
% %\caption{Cost Matrix Computation}
% %\label{tab:Cost03}
% \centering
\begin{footnotesize}
\begin{align*}
\mathcal{M} =&~\text{Set of pre-synaptic neurons of a cluster}\\
\mathcal{N} =&~\text{Set of post-synaptic neurons of a cluster}\\\
\mathcal{S} =&~\text{Set of synapses of a cluster}\\
\eta_{i,k} =&~\text{spikes on the synapse } s_{i,k}\in\mathcal{S}\\
e_{j,l} =&~\text{State transition time of the } (j,l)^\text{th} \text{ OxRRAM cell in a  crossbar} \\
%\mathcal{S}_H =&~\text{State transition time of a crossbar in HRS state} \\
%\mathcal{S}_{L_{1-3}} =&~\text{State transition time of a crossbar in LRS state } L_{1-3} \\
x_{i,j} =& \begin{cases}
1 & \text{if pre-synaptic neuron } m_i\in\mathcal{M} \text{ is mapped to crossbar input } I_j\\
0 & \text{otherwise}
\end{cases}\\
y_{k,l} =& \begin{cases}
1 & \text{if post-synaptic neuron } n_k\in\mathcal{N} \text{ is mapped to crossbar output } O_l\\
0 & \text{otherwise}
\end{cases}
\end{align*}
%\end{small}
% \end{minipage}
% }
% \caption{Notations and legends used in the formulation.}
% \label{tab:notations}
% %\end{small}
% \end{table}
\end{footnotesize}
\normalsize
}

\mr{
Following are the constraints.
\begin{itemize}
    \item A pre-synaptic neuron can be mapped to exactly one input port of a crossbar, i.e.,
    \begin{equation}
        \label{eq:pre-mapping}
        \footnotesize \sum_{\forall j} x_{i,j} = 1~~~~\forall~i
    \end{equation}
    \item A post-synaptic neuron can be mapped to exactly one output port of a crossbar, i.e.,
    \begin{equation}
        \label{eq:post-mapping}
        \footnotesize \sum_{\forall l} y_{k,l} = 1~~~~\forall~k
    \end{equation}
\end{itemize}
}

\mr{
We formulate the optimization problem as follows. \ineq{x_{i,j}\cdot y_{k,l}} defines the mapping of synapse \ineq{s_{i,k}\in\mathcal{S}} to the \ineq{(j,l)^\text{th}} OxRRAM cell in the crossbar. The inference lifetime of this mapping is
\begin{equation}
    \label{eq:inference_lifetime}
    \footnotesize \mathcal{L}_{i,j,k,l} = \frac{e_{j,l}}{\eta_{i,k}}
\end{equation}
}

\mr{
The optimization problem is
\begin{equation}
    \label{eq:optimization_problem}
    \footnotesize \textbf{Maximize}~tRPI = \underset{\forall i,j,k,l}{\texttt{min}}~x_{i,j}\cdot y_{k,l}\cdot\mathcal{L}_{i,j,k,l}
\end{equation}
}

\mr{
The non-linear operation of multiplication of two binary variables \ineq{x_{i,j}} and \ineq{y_{k,l}} is linearized by introducing a new product variable \ineq{z_{i,j,k,l}}, with the following additional constraints.
\begin{itemize}
    \item If \ineq{x_{i,j} = 0} and/or \ineq{y_{k,l} = 0}, then \ineq{z_{i,j,k,l} = 0}, i.e.,
    \begin{equation}
        \label{eq:value_pf_z_0}
        \footnotesize z_{i,j,k,l} \leq x_{i,j} \text{ and }  z_{i,j,k,l} \leq y_{k,l}
    \end{equation}
    \item If \ineq{x_{i,j} = 1} and \ineq{y_{k,l} = 1}, then \ineq{z_{i,j,k,l} = 1}, i.e.,
    \begin{equation}
        \label{eq:value_pf_z_1}
        \footnotesize z_{i,j,k,l} \geq x_{i,j} +  z_{i,j,k,l} - 1
    \end{equation}
\end{itemize}
}

\mr{
The new optimization problem is
\begin{equation}
    \label{eq:new_optimization}
    \footnotesize \textbf{Maximize} \underset{\forall i,j,k,l}{\texttt{min}}~z_{i,j,k,l}\cdot\mathcal{L}_{i,j,k,l}
\end{equation}
}

\mr{
This max-min optimization problem is a convex one (proof of KKT conditions are omitted for space limitations).
The problem can be solved using CVXPY by introducing a slack variable \ineq{\tau} as
\begin{equation}
    \label{eq:max_min}
    \footnotesize \textbf{Maximize } \tau~\ni \tau \leq z_{i,j,k,l}\cdot\mathcal{L}_{i,j,k,l}~\forall~i,j,k,l
\end{equation}
}

\mr{
To incorporate the criticality of a synaptic connection (see Section~\ref{sec:accuracy}), we assign a very small number as the spike count \ineq{\eta} for the synapse. In other words, the spike count of critical synapses is identified using the training set, while those for non-critical synapses are set to a very small value. In this way, we force \ineq{\mathcal{L}} (see Eq.~\ref{eq:inference_lifetime}) to a very large value for the non-critical synapses. This allows the convex optimizer to eliminate them from the critical path of determining the reprogramming interval \ineq{tRPI} (see Eq.~\ref{eq:max_min}).
}

%% file: sections/evaluation.tex
\mr{
We evaluate the proposed design methodology for OxRRAM-based neuromorphic hardware. }
%Our simulation framework includes a cycle-level in-house neuromorphic simulator~\cite{neuroxplorer}. 
We configure NeuroXplorer with the hardware parameters listed in Table~\ref{tab:hw_parameters}.

%\vspace{-10pt}
\begin{table}[h!]
    \caption{Major simulation parameters extracted from~\cite{mallik2017design}.}
	\label{tab:hw_parameters}
	\vspace{-5pt}
	\centering
	{\fontsize{6}{10}\selectfont
		\begin{tabular}{lp{5cm}}
			\hline
			Neuron technology & 16nm CMOS (original design is at 14nm FinFET)\\
			\hline
			Synapse technology & {HfO${}_2$-based OxRRAM}~\cite{mallik2017design}\\
			\hline
			Supply voltage & 1.0V\\
			\hline
			Energy per spike & 23.6pJ at 30Hz spike frequency\\
			\hline
			Energy per routing & 3pJ\\
			\hline
			Switch bandwidth & 3.44 G. Events/s\\
			\hline
% 			NVM related & 1T-1R\\
% 			& PCM cell SET: 24 cycles\\
% 			& PCM cell RESET: 18 cycles\\
% 			& Program \& Verify: 35 cycles\\
% 			\hline
	\end{tabular}}
\end{table}

%Circuit-level simulations are performed with technology parameters from the predictive technology model (PTM)~\cite{zhao2007predictive} and RRAM-specific parameters from~\cite{chen2015compact}.

\mr{
We use five commonly-used convolutional neural network (CNN) applications with 2-bit quantized synaptic weights. These applications are described in Table~\ref{tab:apps}.
}

%\vspace{-10pt}
\begin{table}[h!]
	\renewcommand{\arraystretch}{0.8}
	\setlength{\tabcolsep}{2pt}
	\caption{\mr{CNN applications used to evaluate the proposed design.}}% The (top-1) accuracy with 2-bit quantization is significantly lower than with full-precision as reported in prior works~\cite{krishnamoorthi2018quantizing}.}}
	\label{tab:apps}
	\vspace{-5pt}
	\centering
	\begin{threeparttable}
	{\fontsize{6}{10}\selectfont
	    %\vspace{-10pt}
		\begin{tabular}{cc|ccc|cc}
			\hline
			 \textbf{CNN} &
			\textbf{Dataset} &
			\textbf{Neurons} & \textbf{Synapses} & \textbf{Avg. Spikes/Image} & \textbf{Accuracy-Full} & \textbf{Accuracy-2 bit}\\
			\hline
			LeNet & CIFAR-10 & 80,271 & 275,110 & 724,565 & 86.3\% & 55.4\%\\
			AlexNet & CIFAR-10 & 127,894 & 3,873,222 & 7,055,109 & 66.4\% & 69.5\%\\
			VGGNet & CIFAR-10 & 448,484 & 22,215,209 & 12,826,673 & 81.4 \% & 55.3\%\\
			ResNet & CIFAR-10 & 266,799 & 5,391,616 & 7,339,322 & 57.4\% & 48.0\%\\
			DenseNet & CIFAR-10 & 365,200 & 11,198,470 & 1,250,976 & 46.3\% & 28.2\%\\
			\hline
	\end{tabular}}
	\end{threeparttable}
\end{table}

We evaluate the following techniques.
\begin{itemize}
    \item \mr{\textit{\underline{\sm{}}}. This Baseline approach first clusters a machine-learning inference model to minimize the inter-cluster spike communication~\cite{spinemap}. Clusters are then mapped to crossbars of a neuromorphic hardware with synapses of each cluster implemented randomly on OxRRAM cells of a crossbar.}
    
    \item \mr{\textit{\underline{Endurer}}. This is our previous work, which addresses the reprogramming of model parameters on crossbars of a neuromorphic hardware to maintain model integrity~\cite{song2021improving}. A machine learning model is clustered using \sm{}. Clusters are placed to crossbars to maximize inference lifetime. To map cluster synapses to the OxRRAM cells of a crossbar, Endurer uses a binary non-linear optimization problem formulation.}
    
    \item \mr{\textit{\underline{Proposed}}. The proposed approach is based on Endurer. It introduces the following two new changes to Endurer -- 1) it characterizes a machine learning model to identify non-critical synapses such that they could be eliminated from the critical path of determining the reprogramming interval, and 2) the convex optimization formulation and the proposed linearization technique improves the solution quality and improves the speed-up, accelerating the design space exploration.}
    
    %\item \textit{\underline{Proposed}}. This is proposed framework which reduces the reprogramming-related system overhead by intelligent model partitioning and cluster placement.
\end{itemize}

\subsection{Accuracy}
\begin{figure}[h!]
	\centering
	\vspace{-5pt}
	\centerline{\includegraphics[width=0.99\columnwidth]{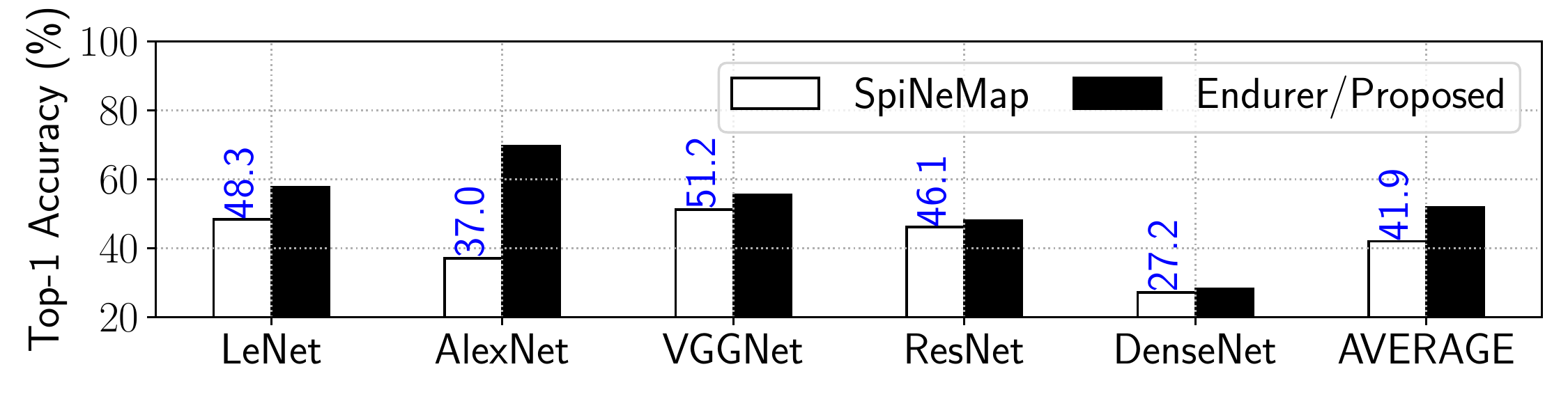}}
	\vspace{-10pt}
	%\caption{An example of spiking neural network.}
	\caption{\mr{Accuracy improvement due to periodic reprogramming.}}
	\vspace{-5pt}
	\label{fig:drop}
\end{figure}

\mr{
Figure~\ref{fig:drop} reports the accuracy improvement due to periodic reprogramming of model parameters in Endurer and the proposed approach compared to \sm{}, where no reprogramming is performed. We observe that by enabling reprogramming of model parameters, model accuracy can be improved by 25\% (between 3\% and 87\%). This is because, without periodic reprogramming in place, model parameters may drift due to frequent accesses of OxRRAM cells where these parameters are programmed. Parameter drift leads to lower accuracy. Additionally, the extent of accuracy impact depends on the specific model that is programmed to the hardware. For AlexNet, we observe a 47\% drop, while for ResNet and DenseNet, the drop is only 4\%.
}

\subsection{System Overhead}
\begin{figure}[h!]
	\centering
	\vspace{-5pt}
	\centerline{\includegraphics[width=0.89\columnwidth]{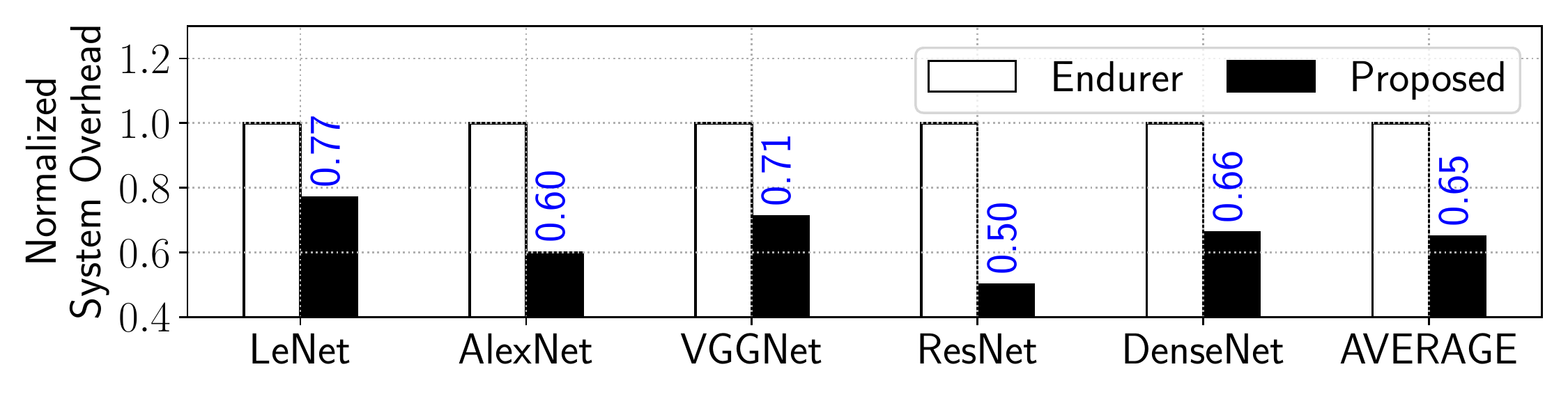}}
	\vspace{-10pt}
	%\caption{An example of spiking neural network.}
	\caption{\mr{Overhead improvement of the proposed approach.}}
	\vspace{-5pt}
	\label{fig:overhead}
\end{figure}

\mr{
Figure~\ref{fig:overhead} reports the system overhead of the proposed approach compared to Endurer for the evaluated CNNs. Results are normalized to Endurer. Since \sm{} does not involve periodic reprogramming, so there is no system overhead. We have therefore not shown \sm{} in the figure. 
}

\mr{
We observe that the system overhead of the proposed approach is on average 35\% lower than Endurer (between 23\% and 50\%). This improvement is due to the increase of reprogramming interval in the proposed approach. Such improvement is attributed to two factors. First, non-critical synapses are not on the critical path for determining the reprogramming interval \ineq{tRPI} in the proposed approach, while such synapses are factored in determining \ineq{tRPI} in Endurer. Second, the convex optimizer CVXPY of the proposed approach generates a better solution than the approximate binary non-linear optimization technique of Endurer.
}

\mr{
We also observe that the improvement is usually higher for models with higher fraction of non-critical synapses. Therefore, the improvement for ResNet is higher than LeNet.
}

%\subsection{Solution Time}

% Figure~\ref{fig:overhead} plots the system overhead (in arbitrary units) incurred in periodically reprogramming the synaptic weights to the neuromorphic hardware. We report results for three state-of-the-art implementation approaches. We make the following three key observations.

% \begin{figure}[h!]
% 	\centering
% 	\vspace{-5pt}
% 	\centerline{\includegraphics[width=0.89\columnwidth]{dt2021/images/overhead_variation.pdf}}
% 	\vspace{-10pt}
% 	%\caption{An example of spiking neural network.}
% 	\caption{System overhead in mitigating read disturbance-related reliability issues of neuromorphic hardware. We compare three state-of-the-art implementation approaches.}
% 	\vspace{-5pt}
% 	\label{fig:overhead}
% \end{figure}

% First, the system overhead is higher for complex models such as VGG, compared to a simple model like LeNet. This is because complex and large models have higher spikes on their synapses for each inference input. Therefore, the OxRRAM on which these models are implemented have lower tRPI, which leads to frequent reprogramming. Second, the system overhead of PyCARL is higher than PSOPART. This is because PyCARL explicitly maximizes each crossbar utilization. Therefore, the number of spikes in each crossbar is higher, which leads to a lower tRPI. Finally, EANC is consistently better than both the techniques. This is because to minimize energy of crossbars, EANC reduces the number of spikes generated in each cluster, which increases the tRPI.

%% file: sections/conclusions.tex
In this work, we study the resistance drift related reliability issues in OxRRAM-based neuromorphic hardware used to implement machine learning inference models. Through circuit-level simulations we show the dependence of these issues on 1) the resistance state of an OxRRAM cell (model parameter dependency), 2) the current through the cell (circuit dependency), and 3) the spikes propagating through the cell (workload dependency). We incorporate this study in a system software framework and show a significant accuracy drop due to resistance drift.
To maintain the integrity of machine learning inference, model parameters need to be reprogrammed to the hardware periodically, which incurs a significant system overhead.
We propose an approach to minimizing this system overhead by first analyzing a machine learning model to identify non-critical synapses, and then proposing a convex optimization solution to maximize the reprogram interval. The proposed optimizer eliminates the non-critical synapses from the critical path of determining the reprogram interval. Evaluations with five commonly-used CNN applications show an average 35\% improvement in the system overhead. 

%% file: sections/ack.tex
This work is supported by the National Science Foundation Faculty Early Career Development Award CCF-1942697 (CAREER: Facilitating Dependable Neuromorphic Computing: Vision, Architecture, and Impact on Programmability).

%% file: sections/bio.tex
\begin{IEEEbiographynophoto}{Ankita Paul}
Ankita Paul is currently pursuing a Ph.D. degree from Drexel University under the supervision of Dr. Anup Das. She received a Bachelor's degree from West Bengal University of Technology in 2016. Her research interests include brain inspired computing, deep learning, and machine learning.
\end{IEEEbiographynophoto}

\begin{IEEEbiographynophoto}{Shihao Song}
Shihao Song is currently pursuing a Ph.D. degree from Drexel University under the supervision of Dr. Anup Das. He received a Bachelor's degree from Drexel University in 2017. His research interests include computer architecture, non-volatile memory, and compiler design for neuromorphic hardware and accelerators.
\end{IEEEbiographynophoto}

\begin{IEEEbiographynophoto}{Twisha Titirsha}
Twisha Titirsha is currently pursuing a Ph.D. degree from the Department of Electrical and Computer Engineering, Drexel University, Philadelphia. She received a Bachelor's degree from Military Institute of Science and Technology, Bangladesh in 2015. Her research interests include computer architecture, non-volatile memory and mixed-signal circuit design.
\end{IEEEbiographynophoto}

\begin{IEEEbiographynophoto}{Anup Das}
Dr. Anup Das is an Assistant Professor at Drexel University. He received a Ph.D. in Embedded Systems from National University of Singapore in 2014.  Following his Ph.D., he was a post-doctoral fellow at the University of Southampton and a researcher at IMEC. His research focuses on neuromorphic computing and architectural exploration. He is a senior member of the IEEE.
\end{IEEEbiographynophoto}